\documentclass[conference]{IEEEtran}
\IEEEoverridecommandlockouts
\usepackage{multirow}
\usepackage{amsmath,amssymb,amsfonts}
\usepackage{algorithmic}
\usepackage{graphicx}
\usepackage{textcomp}
\usepackage{xcolor}
\usepackage{algorithm}
\usepackage{algorithmic}
\usepackage{amsmath}
\usepackage{amssymb}
\usepackage{booktabs}
\usepackage{graphicx}
\usepackage{nicefrac}
\usepackage{nicefrac}       
\usepackage{amsmath}        
\usepackage{amssymb}        
\usepackage{graphicx}      
\usepackage{booktabs}       
\usepackage{hyperref}       
\def\BibTeX{{\rm B\kern-.05em{\sc i\kern-.025em b}\kern-.08em
    T\kern-.1667em\lower.7ex\hbox{E}\kern-.125emX}}
\begin{document}

\title{Synergy-Aware Residual Flow Matching for Single-Cell Perturbation Prediction}

\author{
\IEEEauthorblockN{Jingrui Wu}
\IEEEauthorblockA{
\textit{Department of Statistics}\\
\textit{Rice University}\\
Houston, TX, USA
}
\and
\IEEEauthorblockN{Wangyun Dan}
\IEEEauthorblockA{
\textit{Department of Biostatistics and Bioinformatics}\\
\textit{Emory University}\\
Atlanta, GA, USA
}
}

\IEEEaftertitletext{\vspace{-0.6cm}}

\maketitle

\begin{abstract}
Predicting single-cell responses to genetic and chemical perturbations is important for understanding regulatory mechanisms and prioritizing therapeutic interventions. Existing methods have made progress by modeling perturbation-conditioned expression shifts with neural networks, graph priors, foundation-model representations, or distributional generative models. However, combinatorial perturbation prediction remains challenging because much of the response can be explained by additive single-perturbation effects, while the biologically informative signal often lies in a smaller non-additive residual induced by perturbation interactions. Directly generating the full post-perturbation expression profile may therefore allocate substantial capacity to the dominant additive component and underemphasize synergy-specific structure. To address this issue, we propose \textbf{SynGeneFlow}, a synergy-aware flow matching framework that anchors the source distribution at an additive prior and learns a residual flow toward the true perturbed state. SynGeneFlow further incorporates perturbation-conditioned gene interaction modeling, where a static gene-gene mask provides biological structure and a perturbation-dependent deformation bias dynamically modulates gene-gene coupling. To preserve population-level response structure, we align both endpoint distributions and biological residual distributions with MMD-based regularization. We evaluate SynGeneFlow on genetic and drug perturbation benchmarks under observed-single, held-out, and cross-modality settings. The results demonstrate that residualizing flow matching around additive perturbation effects provides a biologically grounded and robust framework for single-cell perturbation response prediction.
\end{abstract}

\begin{IEEEkeywords}
Single-cell perturbation prediction, flow matching, combinatorial perturbation synergy.
\end{IEEEkeywords}

\section{Introduction}

Predicting cellular responses to genetic or chemical perturbations at single-cell resolution is a central problem in functional genomics, drug discovery, and therapeutic design~\cite{bunne2023learning,lotfollahi2023predicting}. Accurate \textit{in silico} perturbation models can help uncover regulatory mechanisms, identify candidate intervention targets, and reduce the cost of large-scale experimental screening~\cite{qi2024predicting,vinas2025systema}. This need becomes particularly urgent for combinatorial perturbations, where the number of possible gene or drug combinations grows rapidly as the perturbation space expands~\cite{roohani2024predicting,mathur2022combi}. Since exhaustive profiling of all combinations is experimentally infeasible, a useful model must generalize from observed perturbations to unseen combinations while preserving the biological structure of cellular responses.

A key difficulty is that the response to multiple perturbations is not simply the sum of individual effects. Recent benchmarks show that simple mean-based or additive baselines can perform surprisingly well against more complex neural models~\cite{ahlmann2025deep,csendes2025benchmarking,vinas2025systema}. This suggests that a large part of the observed response can be approximated by additive components inherited from single perturbations. However, the scientifically important signal often lies in the remaining non-additive residual, which reflects interactions between perturbations and is commonly interpreted as combinatorial synergy. Such synergy is crucial for understanding genetic interactions, drug combinations, and context-dependent regulatory mechanisms~\cite{norman2019exploring,roohani2024predicting}. Therefore, a model that only reconstructs the full perturbed expression profile may spend most of its capacity on the dominant additive signal while underrepresenting the smaller but more informative non-additive response.

Single-cell perturbation data further make this problem challenging. Because RNA sequencing is destructive, the same cell cannot be observed before and after perturbation, making the data inherently unpaired and preventing direct cell-level supervision. In addition, single-cell expression profiles are sparse, zero-inflated, and affected by technical noise. Dropout can create missing values that do not correspond to true biological absence~\cite{dai2024gene}, while batch effects and sequencing-depth variation can distort gene-level dependencies~\cite{zhou2025quantifying}. These issues make it difficult to distinguish true perturbation-induced changes from technical artifacts. Moreover, perturbation effects are nonlinear and context-dependent~\cite{xing2025gperturb,song2025decoding}. A static gene co-expression graph can provide useful biological prior knowledge, but it only describes an average dependency pattern. Under different perturbation contexts, the effective coupling between the same pair of genes may be strengthened, weakened, or redirected. Models that treat genes as independent features, or rely only on fixed gene-gene priors, may therefore fail to capture the dynamic regulatory rewiring that gives rise to non-additive combinatorial responses.

We propose \textbf{SynGeneFlow}, a synergy-aware generative framework for single-cell perturbation prediction based on conditional flow matching. The central idea of SynGeneFlow is to model non-additive synergy explicitly rather than treating it as an implicit by-product of full-profile generation. At the representation level, SynGeneFlow introduces a perturbation-conditioned cell encoder. Each gene is represented as a token, and a gene-gene co-expression mask restricts attention to biologically plausible interactions. Instead of using this mask as a fixed graph, the encoder uses the perturbation embedding to generate a dynamic deformation bias for each allowed gene pair. This bias modulates gene-gene attention under the current perturbation, allowing the same gene network to produce different cell representations across perturbation contexts. In this way, SynGeneFlow preserves the denoising benefit of biological structure while capturing perturbation-dependent gene network rewiring.

At the generative level, SynGeneFlow introduces synergy-decomposed flow matching. Rather than transporting noise directly to the full perturbed expression profile, we first construct an additive prior from the control expression and estimated single-perturbation effects. The flow then starts from this additive prediction and learns the residual between the additive prior and the true perturbed expression, which corresponds to the non-additive synergy that cannot be explained by independent perturbation effects. To preserve distributional fidelity, we combine the flow matching objective~\cite{lipman2022flow} with Maximum Mean Discrepancy~\cite{gretton2006kernel}. Beyond aligning the final generated expression distribution, we further align the predicted and true synergy-residual distributions, encouraging the model to capture population-level statistics of the non-additive response. We evaluate SynGeneFlow on the Norman combinatorial genetic perturbation dataset~\cite{norman2019exploring} and the ComboSciPlex drug perturbation dataset~\cite{mathur2022combi}. Across additive and holdout generalization settings, SynGeneFlow improves the accuracy and robustness of \textit{in silico} perturbation prediction, demonstrating the benefit of jointly modeling perturbation-conditioned gene rewiring and synergy-residual distributions.

\section{Related Work}

\subsection{Foundation Models for Single-Cell Biology.}
Large-scale pretraining has become an increasingly important direction for learning transferable representations of single-cell transcriptomes. Instead of training task-specific models from scratch, foundation-style models aim to learn reusable cell and gene embeddings from large collections of expression profiles. Representative examples include Geneformer, scBERT, scGPT, UCE, and related transformer-based single-cell models~\cite{theodoris2023transfer,yang2022scbert,cui2024scgpt,rosen2023universal,khan2023reusability}. These models typically use masked prediction, tokenized gene representations, or large-scale self-supervised objectives to capture broad transcriptomic structure. As a result, they have shown promise in tasks such as cell-type annotation, representation learning, transfer learning, and biological network prediction.

Despite their broad applicability, the role of foundation models in perturbation-response prediction remains less settled. Most of these models are pretrained on unperturbed or weakly conditioned expression profiles, where the learning objective mainly encourages the recovery of general transcriptomic patterns rather than intervention-specific changes. This mismatch becomes important when the downstream task requires predicting how a cell actively moves from one state to another after a genetic or chemical perturbation. Recent evaluations suggest that foundation models do not consistently outperform simpler baselines in post-perturbation prediction, especially in zero-shot or distribution-shifted settings~\cite{ahlmann2025deep,csendes2025benchmarking,yu2026scdfm}. Broader efforts toward virtual cells and context-aware perturbation models further highlight that transferable representations alone are insufficient unless the model also captures the causal or functional effect of interventions~\cite{roohani2025virtual,adduri2025predicting}. These observations motivate methods that go beyond generic cell embeddings and explicitly model perturbation-conditioned response dynamics.

\subsection{Models Tailored to Perturbation Effects.}
A second line of work directly targets perturbation-response prediction by incorporating perturbation identities, biological priors, uncertainty modeling, or generative transition mechanisms. CPA models perturbation composition by embedding drugs, genes, dosages, and cellular contexts in a shared latent space, enabling extrapolation to unseen combinations through learned condition embeddings~\cite{lotfollahi2023predicting}. GEARS instead injects biological structure into the architecture by using gene-gene prior knowledge, including co-expression and Gene Ontology relationships, to improve generalization to unseen genetic perturbations~\cite{roohani2024predicting}. PerturbNet and GPerturb further explore probabilistic formulations for predicting responses under unseen chemical or genetic conditions, aiming to better quantify uncertainty and distributional variation~\cite{yu2025perturbnet,xing2025gperturb}. Other works study large-scale chemical or combinatorial perturbation profiling, providing important datasets and benchmarks for evaluating whether models can generalize beyond observed interventions~\cite{srivatsan2020massively,mathur2022combi,qi2024predicting}.

More recently, generative models have been used to describe perturbation response as a transition between cellular distributions. Methods such as scDiffusion, CellFlow, and UNLASTING formulate this problem using diffusion, flow matching, or bridge-based generative modeling, allowing the model to learn continuous transformations from control states to perturbed states~\cite{bunne2023learning,klein2025cellflow,lipman2022flow,xi2026grounding, xi2026medlvr}. These approaches move beyond deterministic mean prediction and are better aligned with the unpaired and distributional nature of single-cell perturbation data. At the same time, recent studies have emphasized that evaluation should not only reward mean recovery, since perturbations can change variance, heterogeneity, and subpopulation structure~\cite{mejia2025diversity,ramakrishnan2025modeling,vinas2025systema, xi2026radiographic}. In combinatorial settings, an additional difficulty is that many methods compose perturbations implicitly or approximately additively. Such a design can explain the dominant response but may underemphasize the smaller non-additive component produced by interactions between perturbations. SynGeneFlow is designed around this gap: it dynamically modulates gene-gene interactions according to the perturbation context and uses an additive prior only as a starting point, forcing the flow model to focus on the synergy residual rather than reconstructing the entire response from scratch.

\section{Method}

We introduce \textbf{SynGeneFlow}, a synergy-aware flow matching framework for single-cell perturbation response prediction. Let $\mathcal{G}=\{g_1,\ldots,g_G\}$ denote the set of profiled genes. A control cell is represented by its expression vector $c_x\in\mathbb{R}_+^G$, and the corresponding post-perturbation expression is denoted by $x_1\in\mathbb{R}_+^G$. The perturbation condition is encoded as a multi-hot vector $c_p\in\{0,1\}^{d_p}$, where $c_p[j]=1$ indicates that the $j$-th genetic or chemical perturbation is applied. We denote the active perturbation set by $\operatorname{supp}(c_p)=\{j:c_p[j]=1\}$. Because single-cell perturbation data are unpaired, we do not assume that $c_x$ and $x_1$ come from the same cell. For each condition, we sample a control cell $c_x \sim q_0(\cdot)$ and an independently observed perturbed cell $x_1 \sim q_1(\cdot\mid c_p)$. The resulting tuple $(c_x,c_p,x_1)$ is used for conditional distribution matching, with the objective of learning $p_\theta(x_1\mid c_x,c_p)$ to capture the population-level distribution of perturbed cells and generalize to unseen perturbation combinations.

SynGeneFlow is motivated by the structure of combinatorial perturbation responses: a large fraction of the response can often be explained by additive single-perturbation effects, while the biologically informative signal lies in the non-additive residual induced by perturbation interactions. Instead of transporting a generic noise source to the full perturbed profile, SynGeneFlow anchors the source distribution at an additive prior and learns a residual flow toward the true perturbed state. The framework further couples this residualized generative process with perturbation-conditioned gene interaction modeling, where a static co-expression mask defines biologically plausible edges and a perturbation-dependent deformation bias modulates their strengths. This design links dynamic gene-network rewiring at the representation level with synergy-residual modeling at the generative level.

\begin{figure*}[t]
    \centering
    \includegraphics[width=\textwidth]{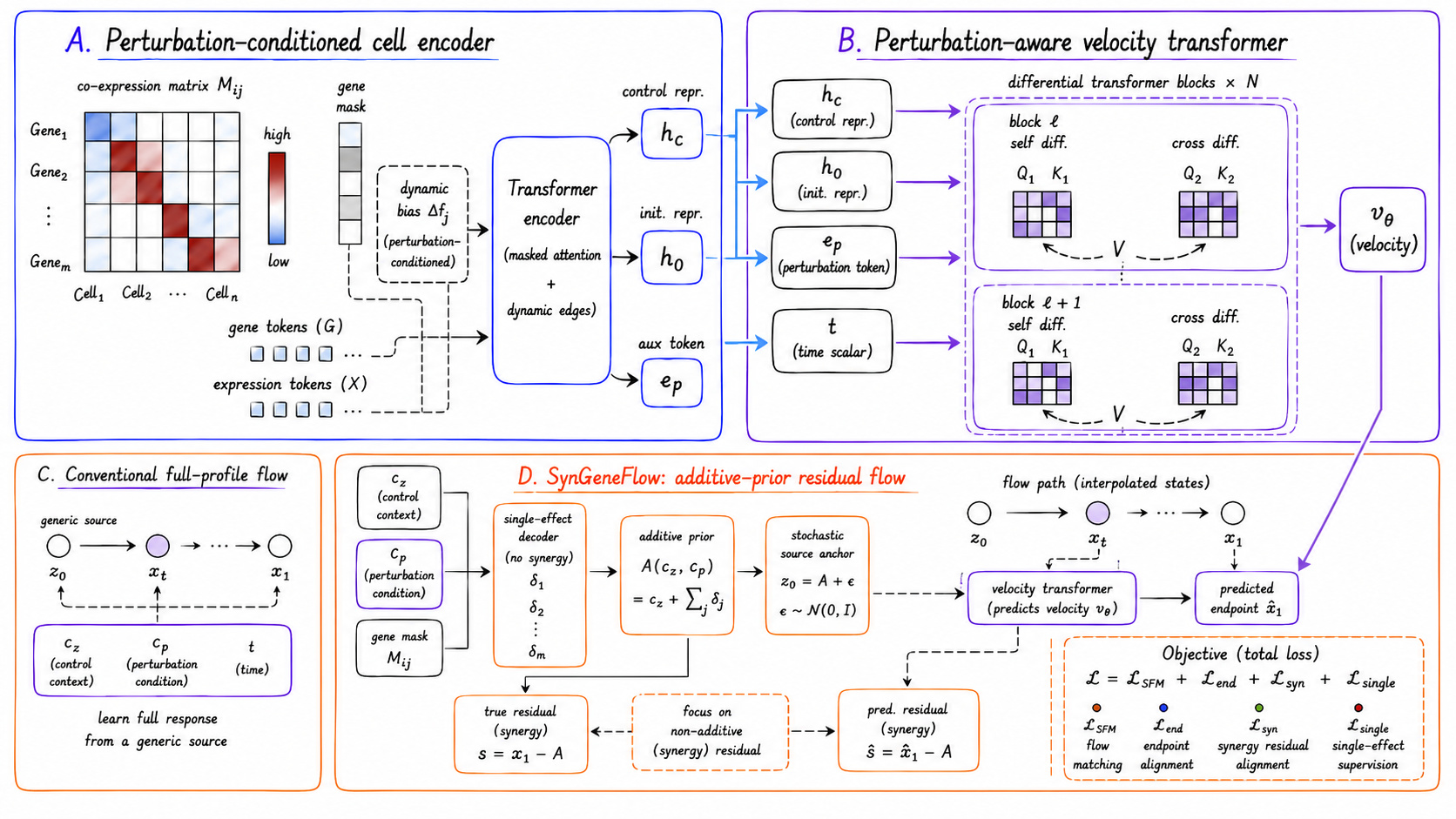}
    \caption{
    \textbf{Overview of SynGeneFlow.}
    \textbf{A}, Perturbation-conditioned cell encoder.
    \textbf{B}, Perturbation-aware velocity Transformer.
    \textbf{C}, Conventional full-profile flow matching.
    \textbf{D}, Additive-prior residual flow matching for non-additive perturbation effects.
    }
    \label{fig:method_overview}
\end{figure*}

\subsection{Synergy-Residual Flow Matching}

Flow matching learns a time-dependent velocity field that transports samples from a source distribution to a target distribution along a continuous path~\cite{lipman2022flow}. In standard conditional flow matching, the source distribution is often chosen as random noise or another generic base distribution. For combinatorial perturbation prediction, this forces the model to reconstruct the full post-perturbation expression profile, including the dominant additive component. We instead define a biologically anchored source distribution centered at an additive prediction, so that the learned flow focuses on the non-additive response.

We first define an additive operator
\begin{equation}
\mathcal{A}(c_x,c_p)
=
c_x+
\sum_{j\in\operatorname{supp}(c_p)}
\delta_j,
\qquad
\delta_j=g_\phi(e_{p_j},c_x),
\label{eq:additive_prior}
\end{equation}
where $e_{p_j}\in\mathbb{R}^{d_p'}$ is the embedding of the $j$-th perturbation, $d_p'$ is the perturbation embedding dimension, and $g_\phi$ is a learnable effect decoder that predicts the single-perturbation shift $\delta_j\in\mathbb{R}^{G}$ conditioned on the control state. For a multi-perturbation condition, $\mathcal{A}(c_x,c_p)$ sums the estimated effects of all active perturbations and represents the response explained by independent perturbation effects. When single-perturbation profiles are unavailable, the corresponding effect is extrapolated from the perturbation embedding through $g_\phi$.

To retain within-condition variability, we add a small control-anchored noise term to the additive prior:
\begin{equation}
z_0
=
\mathcal{A}(c_x,c_p)+\epsilon,
\qquad
\epsilon\sim q_\tau(\epsilon\mid c_x),
\label{eq:stochastic_anchor}
\end{equation}
where $z_0$ is the stochastic additive anchor, $q_\tau$ is a control-anchored source noise distribution, and $\tau$ controls the noise magnitude. In our implementation, $q_\tau$ is a diagonal Gaussian whose scale is estimated from training control-cell statistics and fixed at validation and test time. The case $\tau=0$ recovers the deterministic additive source. Randomness is injected only at the source anchor, while the ODE trajectory conditioned on a sampled $z_0$ remains deterministic.

Given a real perturbed expression $x_1\sim q_1(\cdot\mid c_x,c_p)$, we define the interpolation path
\begin{equation}
x_t
=
(1-t)z_0+t x_1,
\qquad
t\sim\mathcal{U}(0,1).
\label{eq:synergy_path}
\end{equation}
Under this path, the reference velocity is
\begin{equation}
v^\star
=
x_1-z_0
=
x_1-\mathcal{A}(c_x,c_p)-\epsilon
\triangleq s_\tau.
\label{eq:stochastic_velocity_target}
\end{equation}
Here $s_\tau$ is the stochastic transport residual used for velocity supervision. We separately define the biological synergy residual as
\begin{equation}
s
=
x_1-\mathcal{A}(c_x,c_p),
\label{eq:biological_synergy_residual}
\end{equation}
which removes the additive component and represents the non-additive perturbation response. The velocity field $v_\theta(x_t,t,c_x,c_p)$ is trained to predict $s_\tau$, while the residual distribution alignment in Section~\ref{sec:dual_alignment} is applied to the biological residual $s$.

The corresponding continuous dynamics are
\begin{equation}
\frac{dX_t}{dt}
=
v_\theta(X_t,t,c_x,c_p),
\label{eq:ode}
\end{equation}
where $X_t$ denotes the generated expression state at time $t$. The synergy-residual flow matching loss is
\begin{equation}
\mathcal{L}_{\mathrm{SFM}}
=
\mathbb{E}
\left[
\left\|
v_\theta(x_t,t,c_x,c_p)-s_\tau
\right\|_2^2
\right],
\label{eq:sfm_loss}
\end{equation}
where the expectation is taken over $(c_x,c_p,x_1)$, $t\sim\mathcal{U}(0,1)$, and $\epsilon\sim q_\tau(\epsilon\mid c_x)$. Since the source state is already close to the target under the additive prior, the transport path is shorter and the model capacity is directed toward the interaction-induced residual.

To prevent the additive operator from absorbing non-additive effects, we impose a single-perturbation consistency loss:
\begin{equation}
\mathcal{L}_{\mathrm{single}}
=
\mathbb{E}_{a\in\mathcal{P}_{\mathrm{single}}^{\mathrm{train}}}
\left[
\left\|
\mathcal{A}(c_x,\{a\})-x_1^{(a)}
\right\|_2^2
\right],
\label{eq:single_loss}
\end{equation}
where $\mathcal{P}_{\mathrm{single}}^{\mathrm{train}}$ denotes the set of single perturbations observed in the training set, and $x_1^{(a)}$ is the observed expression under the single perturbation $a$. This term encourages $\mathcal{A}$ to learn additive single-perturbation effects, leaving higher-order interactions to the residual velocity field.

\subsection{Dual Distribution Alignment}
\label{sec:dual_alignment}

The velocity loss in Eq.~\ref{eq:sfm_loss} supervises sample-wise stochastic transport residuals along the interpolation path, but it does not explicitly constrain the generated population distribution. We therefore introduce a dual distribution alignment objective based on Maximum Mean Discrepancy (MMD)~\cite{gretton2006kernel}. MMD is suitable for single-cell perturbation data because it can be estimated from unpaired mini-batches and does not require cell-level correspondence.

For a sampled intermediate state $x_t$, we form a one-step endpoint approximation
\begin{equation}
\hat{x}_1
=
x_t+(1-t)v_\theta(x_t,t,c_x,c_p),
\label{eq:one_step_endpoint}
\end{equation}
and compute the predicted synergy residual by subtracting the additive prior:
\begin{equation}
\hat{s}
=
\hat{x}_1-\mathcal{A}(c_x,c_p).
\label{eq:predicted_synergy_residual}
\end{equation}
The target residual is the biological residual $s=x_1-\mathcal{A}(c_x,c_p)$.

We align both the endpoint distribution and the biological residual distribution:
\begin{equation}
\begin{aligned}
\mathcal{L}_{\mathrm{MMD}}^{\mathrm{end}}
&=
\mathrm{MMD}^2(\{\hat{x}_1\},\{x_1\}), \\
\mathcal{L}_{\mathrm{MMD}}^{\mathrm{syn}}
&=
\mathrm{MMD}^2(\{\hat{s}\},\{s\}).
\end{aligned}
\label{eq:dual_mmd}
\end{equation}
The endpoint term improves population-level fidelity, while the synergy term explicitly matches the distribution of non-additive residuals. These two terms are complementary: endpoint alignment constrains the final generated cell population, whereas residual alignment encourages the model to preserve the variance and subpopulation structure of the synergy component.

We compute MMD with a multi-kernel Gaussian RBF mixture:
\begin{equation}
k_{\mathrm{mix}}(u,v)
=
\frac{1}{L_m}
\sum_{\ell=1}^{L_m}
\exp\left(
-\frac{\|u-v\|_2^2}{2\sigma_\ell^2}
\right),
\label{eq:mmd_kernel}
\end{equation}
where $L_m$ is the number of kernels and $\sigma_\ell$ is the $\ell$-th bandwidth. Bandwidths are estimated from mini-batch pairwise distances, allowing the kernel scale to adapt to expression profiles or residual vectors.

The final objective is
\begin{equation}
\mathcal{L}
=
\mathcal{L}_{\mathrm{SFM}}
+
\lambda_1\mathcal{L}_{\mathrm{MMD}}^{\mathrm{end}}
+
\lambda_2\mathcal{L}_{\mathrm{MMD}}^{\mathrm{syn}}
+
\lambda_3\mathcal{L}_{\mathrm{single}},
\label{eq:final_loss}
\end{equation}
where $\lambda_1,\lambda_2,\lambda_3>0$ balance endpoint fidelity, synergy-distribution fidelity, and additive single-perturbation consistency.

\subsection{Perturbation-Conditioned Gene Encoding}

SynGeneFlow encodes each cell as a sequence of gene tokens. Because the number of genes can be large, training may be performed on a selected subset $I\subseteq\{1,\ldots,G\}$ with $|I|=N_I\ll G$. The expression vectors restricted to this subset are denoted as $c_x^{(I)}$ and $x_t^{(I)}$. Let $G_I=(g_i)_{i\in I}$ be the ordered gene identity sequence. A gene identity encoder $E_g$ and a scalar value encoder $E_v$ produce aligned token sequences:
\begin{equation}
\begin{aligned}
h_c   &= E_v(c_x^{(I)})+Z_I, \\
h_v^0 &= E_v(x_t^{(I)})+Z_I, \\
Z_I   &= E_g(G_I;M_I,e_p).
\end{aligned}
\label{eq:token_init_subset}
\end{equation}
Here $h_c$ is the control representation, $h_v^0$ is the initial latent representation for the intermediate state, $Z_I\in\mathbb{R}^{N_I\times d}$ denotes contextualized gene identity tokens, $d$ is the hidden dimension, $M_I$ is the subgraph of the full gene-gene mask restricted to the selected genes, and $e_p=\mathrm{Emb}(c_p)$ is the perturbation embedding.

We construct the static gene-gene prior from the training data. For genes $i$ and $j$, the edge weight is defined by the absolute Pearson correlation
\begin{equation}
w_{ij}
=
\left|
\frac{\mathrm{Cov}(x_i,x_j)}
{\sigma(x_i)\sigma(x_j)}
\right|,
\label{eq:pearson_graph}
\end{equation}
where $x_i$ and $x_j$ are expression vectors across cells. A $k$-nearest-neighbor procedure retains the strongest correlated neighbors for each gene and yields a sparse adjacency matrix. We convert this adjacency into an attention mask $M_{ij}$, where allowed edges receive finite values and disallowed edges are masked out during attention.

A static co-expression graph provides a useful biological prior, but it only reflects average dependency structure. Perturbations can change the effective coupling between genes. To capture this context dependence, we introduce a perturbation-conditioned deformation bias for each allowed gene pair. For gene representations $h_i$ and $h_j$, we compute $q_i=W_Qh_i$, $k_j=W_Kh_j$, and $v_j=W_Vh_j$. The perturbation-conditioned attention score is
\begin{equation}
\begin{aligned}
a_{ij}^{p}
&=
\frac{q_i^\top k_j}{\sqrt{d}}
+
M_{ij}
+
\Delta_{ij}^{p}, \\
\Delta_{ij}^{p}
&=
w_{\psi}^{\top}
\sigma\!\left(
W_{\psi}
[
h_i\Vert h_j\Vert e_p
]
+b_{\psi}
\right).
\end{aligned}
\label{eq:dynamic_attention_score}
\end{equation}

Here $a_{ij}^{p}$ is the attention logit from gene $i$ to gene $j$ under perturbation condition $c_p$, $\Delta_{ij}^{p}$ is the perturbation-dependent deformation bias, $\Vert$ denotes concatenation, $\sigma(\cdot)$ is a nonlinear activation, and $W_{\psi}$, $b_{\psi}$, and $w_{\psi}$ are learnable parameters. The mask $M_{ij}$ defines biologically plausible gene interactions, while $\Delta_{ij}^{p}$ dynamically adjusts their strengths under the current perturbation.

The resulting attention weights and structural gene representation are
\begin{equation}
\alpha_{ij}^{p}
=
\mathrm{softmax}_{j}(a_{ij}^{p}),
\qquad
h_i^{\mathrm{gene}}
=
\sum_j \alpha_{ij}^{p}v_j.
\label{eq:gene_attention}
\end{equation}
The final cell token combines this structural representation with the expression-value embedding:
\begin{equation}
h_i^{\mathrm{cell}}
=
h_i^{\mathrm{gene}}+h_i^{\mathrm{expr}},
\label{eq:cell_token}
\end{equation}
where $h_i^{\mathrm{expr}}$ is the expression-value embedding of gene $i$. Thus, the encoder uses the static graph as a biologically plausible interaction space and learns perturbation-specific edge modulation within that space.

\subsection{Perturbation-Aware Velocity Transformer}

The velocity network estimates the stochastic transport residual $s_\tau$ from sparse and noisy single-cell measurements. It takes the intermediate-state tokens $h_v^0$, the control tokens $h_c$, the perturbation embedding $e_p$, and a time embedding as inputs:
\begin{equation}
t_{\mathrm{emb}}
=
\mathrm{MLP}(\mathrm{SinCos}(t)).
\label{eq:time_embedding}
\end{equation}
We use a perturbation-aware Transformer backbone. At layer $\ell$, the latent representation $h_v^\ell$ is first conditioned on the perturbation by
\begin{equation}
\bar{h}_v^\ell
=
\mathrm{MLP}_\ell
\left(
[h_v^\ell\Vert \mathbf{1}_{N_I}\otimes e_p]
\right),
\label{eq:pert_injection}
\end{equation}
where $\mathbf{1}_{N_I}\otimes e_p$ replicates the perturbation embedding across gene tokens. It is then refined by self-attention and cross-attention against the control representation:
\begin{equation}
\begin{aligned}
\tilde{h}_v^\ell
&=
\bar{h}_v^\ell
+
\mathrm{Attn}
(
\bar{h}_v^\ell,
\bar{h}_v^\ell;
t_{\mathrm{emb}}
), \\
h_v^{\ell+1}
&=
\tilde{h}_v^\ell
+
\mathrm{Attn}
(
\tilde{h}_v^\ell,
h_c;
t_{\mathrm{emb}}
).
\end{aligned}
\label{eq:velocity_transformer_layer}
\end{equation}
Here $\mathrm{Attn}(\cdot)$ denotes a time-conditioned attention block. In implementation, it can be instantiated with differential attention to suppress shared noisy attention patterns, but the residual-flow formulation is independent of this backbone choice. After $L_v$ layers, the final representation is decoded into the velocity field:
\begin{equation}
v_\theta^{(I)}
=
D
\left(
[h_v^{L_v}\Vert \mathbf{1}_{N_I}\otimes e_p]
\right),
\label{eq:velocity_decode}
\end{equation}
where $D$ is the velocity decoder and $L_v$ is the number of velocity Transformer layers. This velocity is used in Eq.~\ref{eq:sfm_loss} to predict the stochastic transport residual $s_\tau$.

\subsection{Training and Inference}

During training, each mini-batch contains triplets $(c_x,c_p,x_1)$. When subset training is used, all source, path, residual, and loss terms are computed on the selected gene subset $I$. Specifically, we construct $\mathcal{A}^{(I)}(c_x,c_p)$, sample $\epsilon^{(I)}\sim q_\tau(\epsilon^{(I)}\mid c_x^{(I)})$, form $z_0^{(I)}=\mathcal{A}^{(I)}(c_x,c_p)+\epsilon^{(I)}$, and compute the interpolation state $x_t^{(I)}=(1-t)z_0^{(I)}+t x_1^{(I)}$. The velocity Transformer predicts $v_\theta^{(I)}$, and the endpoint and residual used for distribution alignment are
\begin{equation}
\hat{x}_1^{(I)}
=
x_t^{(I)}+(1-t)v_\theta^{(I)},
\qquad
\hat{s}^{(I)}
=
\hat{x}_1^{(I)}
-
\mathcal{A}^{(I)}(c_x,c_p).
\label{eq:subset_endpoint_residual}
\end{equation}
Endpoint MMD is evaluated on $\hat{x}_1^{(I)}$ and $x_1^{(I)}$, while synergy MMD is evaluated on $\hat{s}^{(I)}$ and $s^{(I)}=x_1^{(I)}-\mathcal{A}^{(I)}(c_x,c_p)$.

At inference time, SynGeneFlow samples the same type of stochastic additive anchor:
\begin{equation}
X_0^{(I)}
=
\mathcal{A}^{(I)}(c_x,c_p)+\epsilon^{(I)},
\qquad
\epsilon^{(I)}\sim q_\tau(\epsilon^{(I)}\mid c_x^{(I)}).
\label{eq:inference_init}
\end{equation}
The scale of $q_\tau$ is estimated from training control-cell statistics and kept fixed at test time. Given a sampled $X_0^{(I)}$, the ODE trajectory is deterministic. We integrate Eq.~\ref{eq:ode} from $t=0$ to $t=1$ using Euler steps:
\begin{equation}
X_{k+1}^{(I)}
=
X_k^{(I)}
+
\Delta t\,
v_\theta(X_k^{(I)},t_k,c_x,c_p),
\qquad
t_k=\frac{k}{K}.
\label{eq:euler_inference}
\end{equation}
The final output $\hat{x}_1^{(I)}=X_K^{(I)}$ is the predicted post-perturbation expression on the selected gene set. For population-level prediction, we propagate a batch of control cells under the same perturbation condition; when $\tau>0$, independent stochastic anchors are sampled to provide additional within-condition variability.

The computational cost of the gene encoder and velocity Transformer attention is $\mathcal{O}(N_I^2)$ per layer. The deformation bias is evaluated only on retained graph edges; with $E_I$ allowed edges and edge-MLP hidden dimension $d_\psi$, its cost is $\mathcal{O}(E_I d_\psi)$. The additive prior requires $|\operatorname{supp}(c_p)|$ effect-decoder evaluations, and the two MMD terms require $\mathcal{O}(B^2)$ pairwise kernel evaluations for mini-batch size $B$.
\begin{table*}[t]
\centering
\caption{Comparison of different methods across evaluation metrics on the Norman additive split.}
\resizebox{\textwidth}{!}{
\begin{tabular}{lcccccccc}
\toprule
\textbf{Model} & \textbf{L2 $\downarrow$} & \textbf{MSE $\downarrow$} & \textbf{MAE $\downarrow$} & \textbf{DE-Spearman $\rho$ $\uparrow$} & \textbf{Pearson $\Delta$ $\uparrow$} & \textbf{DS $\uparrow$} & \textbf{Pearson $\hat{\Delta}$ $\uparrow$} & \textbf{Pearson $\hat{\Delta}_{20}$ $\uparrow$} \\
\midrule
Control & 3.9937 & 0.01839 & 0.03953 & N.A. & N.A. & 0.5135 & -0.1695 & -0.1297 \\
Additive & 1.9395 & 0.00448 & 0.02276 & 0.5564 & \underline{0.9024} & 0.9686 & \underline{0.8584} & \underline{0.9244} \\
scGPT & 3.4112 & 0.01349 & 0.03796 & 1.07e-5 & 0.5304 & 0.5404 & 0.2165 & 0.2414 \\
Geneformer & 1.9132 & 0.00410 & 0.02360 & 0.3741 & 0.7732 & 0.8241 & -0.0078 & 0.2239 \\
GEARS & 3.5531 & 0.01387 & 0.06624 & 0.5624 & 0.7421 & 0.8601 & -0.0089 & 0.2032 \\
CPA & 5.7629 & 0.03435 & 0.07894 & 0.0713 & 0.3845 & 0.6021 & -0.0039 & 0.2254 \\
STATE & 17.3330 & 0.30059 & 0.24705 & 0.5288 & -0.0108 & 0.5135 & -0.0069 & 0.2515 \\
CellFlow & 1.7064 & 0.00392 & 0.02207 & 0.5503 & 0.8678 & 0.9321 & 0.8395 & 0.8988 \\
scDFM & \underline{1.7043} & \underline{0.00315} & \underline{0.02155} & \underline{0.5705} & 0.8853 & \underline{0.9737} & 0.8468 & 0.9260 \\
\midrule
SynGeneFlow (ours) & \textbf{1.6736} & \textbf{0.00296} & \textbf{0.02091} & \textbf{0.5928} & \textbf{0.9086} & \textbf{0.9779} & \textbf{0.8642} & \textbf{0.9347} \\
\bottomrule
\end{tabular}
}
\label{tab:additive}
\end{table*}

\section{Experiments}

We evaluate SynGeneFlow under three perturbation prediction regimes. The first tests whether the model can predict unseen double perturbations when the corresponding single perturbations are observed during training. The second evaluates a harder holdout setting, where selected single perturbations and all combinations involving them are excluded from training. The third tests transfer from genetic perturbations to drug-induced transcriptional responses. Across these settings, we examine whether additive-prior residual flow matching improves reconstruction accuracy and perturbation-response recovery while preserving population-level structure.

\subsection{Evaluation Design}

We use two single-cell perturbation benchmarks. The Norman dataset contains CRISPR-based gene overexpression profiles in K562 cells, including both single and double perturbations~\cite{norman2019exploring}. We evaluate two splits on Norman. In the additive split, test double perturbations are composed of single perturbations observed during training, which tests whether a model can recover non-additive interaction effects beyond simple composition. In the holdout split, selected single perturbations and all combinations involving them are removed from training, creating a more difficult extrapolation setting. We further evaluate on ComboSciPlex to test whether the same framework transfers from genetic perturbations to drug perturbations~\cite{srivatsan2020massively,lotfollahi2023predicting}.

All methods are evaluated under the same preprocessing pipeline, gene set, and train/validation/test splits. For Norman, we follow the same additive and holdout protocols used by the direct flow-matching baseline, and for ComboSciPlex we follow the held-out drug-combination evaluation setting.

We compare SynGeneFlow with statistical baselines, task-specific perturbation models, foundation-model baselines, and generative transition baselines. These include the control response, additive composition, CPA~\cite{lotfollahi2023predicting}, GEARS~\cite{roohani2024predicting}, CellFlow~\cite{klein2025cellflow}, STATE~\cite{adduri2025predicting}, Geneformer~\cite{theodoris2023transfer}, scGPT~\cite{cui2024scgpt}, and scDFM. The comparison with scDFM is particularly important because it separates the effect of the proposed additive-prior residual formulation from the general benefit of using flow matching. Performance is evaluated using reconstruction metrics, including L2, MSE, and MAE; perturbation-response metrics, including Pearson $\Delta$, Pearson $\hat{\Delta}$, Pearson $\hat{\Delta}_{20}$, and DE-Spearman $\rho$; and the Discrimination Score (DS), which measures whether predicted perturbation populations remain separable~\cite{roohani2025virtual}.

\subsection{Observed-Single Combinatorial Prediction}
\label{section:experiment-norman-additive}

We first evaluate the Norman additive split, where the single perturbations composing each test double perturbation are already observed during training. This setting focuses on whether a model can recover non-additive residual effects beyond independent single-perturbation composition.

As shown in Table~\ref{tab:additive}, the additive baseline is highly competitive on Pearson-based perturbation-response metrics, reaching 0.9024 on Pearson $\Delta$, 0.8584 on Pearson $\hat{\Delta}$, and 0.9244 on Pearson $\hat{\Delta}_{20}$. This confirms that a large part of the combinatorial response can be explained by first-order composition. Among neural baselines, CellFlow gives strong reconstruction performance, while GEARS achieves the strongest non-flow baseline DE-Spearman $\rho$. scDFM further improves reconstruction, DE ranking, and distributional separation, indicating that direct flow matching is already a strong baseline for combinatorial perturbation prediction.

SynGeneFlow provides consistent gains over scDFM across all metrics in this setting. It reduces L2 from 1.7043 to 1.6736, MSE from 0.00315 to 0.00296, and MAE from 0.02155 to 0.02091. It also improves DE-Spearman $\rho$ from 0.5705 to 0.5928, Pearson $\Delta$ from 0.8853 to 0.9086, Pearson $\hat{\Delta}$ from 0.8468 to 0.8642, and Pearson $\hat{\Delta}_{20}$ from 0.9260 to 0.9347. DS also increases from 0.9737 to 0.9779. These results suggest that anchoring the flow at an additive prior helps recover perturbation-response directions, DE-gene structure, and population separability beyond direct full-profile flow matching.

\subsection{Held-Out Perturbation Generalization}
\label{section:experiment-norman-holdout}

We next evaluate the Norman holdout split, where selected single perturbations and all double perturbations involving them are excluded from training. This setting is substantially more challenging than the additive split because the model must extrapolate both unseen single-perturbation effects and their combinatorial interactions.

Table~\ref{tab:norman_holdout_metrics} reports results for held-out single and double perturbations. For held-out single perturbations, SynGeneFlow achieves the best performance on all metrics except DS. Compared with the strongest direct flow-matching baseline scDFM, SynGeneFlow reduces L2 from 1.6186 to 1.5482, MSE from 0.0030 to 0.0026, and MAE from 0.0190 to 0.0182. It also improves DE-Spearman $\rho$ from 0.6957 to 0.7248, Pearson $\Delta$ from 0.7127 to 0.7426, Pearson $\hat{\Delta}$ from 0.6659 to 0.7045, and Pearson $\hat{\Delta}_{20}$ from 0.8116 to 0.8462. scDFM remains slightly stronger on DS, suggesting that direct flow matching still preserves population-level separability well under extrapolation, while SynGeneFlow more effectively recovers perturbation-response structure.

For held-out double perturbations, the advantage of SynGeneFlow becomes more pronounced. SynGeneFlow again outperforms scDFM on all metrics except DS, reducing L2 from 2.0309 to 1.8846, MSE from 0.0047 to 0.0041, and MAE from 0.0235 to 0.0221. It also improves DE-Spearman $\rho$ from 0.5676 to 0.6312, Pearson $\Delta$ from 0.8357 to 0.8624, Pearson $\hat{\Delta}$ from 0.7769 to 0.8136, and Pearson $\hat{\Delta}_{20}$ from 0.8688 to 0.8965. Although scDFM still attains the highest DS, SynGeneFlow shows consistently stronger reconstruction and response-correlation performance in the hardest extrapolation regime. Overall, these results show that residualizing the transport around additive effects substantially improves held-out perturbation generalization while maintaining competitive population-level structure.

\begin{table*}[t]
\centering
\caption{
\textbf{Performance on the Norman holdout split.}
Results are reported for held-out single and double perturbations across reconstruction, response-correlation, and distributional metrics.
Best results are shown in bold and second-best results are underlined.
}
\label{tab:norman_holdout_metrics}
\setlength{\tabcolsep}{4pt}
\resizebox{\textwidth}{!}{%
\begin{tabular}{llcccccccc}
\toprule
\textbf{Setting}
& \textbf{Model}
& \textbf{L2 $\downarrow$}
& \textbf{MSE $\downarrow$}
& \textbf{MAE $\downarrow$}
& \textbf{DE-Spearman $\rho$ $\uparrow$}
& \textbf{Pearson $\Delta$ $\uparrow$}
& \textbf{DS $\uparrow$}
& \textbf{Pearson $\hat{\Delta}$ $\uparrow$}
& \textbf{Pearson $\hat{\Delta}_{20}$ $\uparrow$} \\
\midrule
Single & Control & 2.6834 & 0.0095 & 0.0263 & N.A. & N.A. & 0.5217 & 0.1618 & 0.1982 \\
Single & scGPT & 2.5007 & 0.0080 & 0.0259 & -0.1139 & 0.4503 & 0.5680 & 0.0747 & 0.0798 \\
Single & GEARS & 2.5641 & 0.0075 & 0.0466 & 0.3569 & 0.6646 & 0.8271 & 0.6356 & 0.7914 \\
Single & Geneformer & 1.6962 & 0.0036 & 0.0191 & 0.3669 & 0.6955 & 0.8070 & 0.5620 & 0.6513 \\
Single & CPA & 5.8060 & 0.0356 & 0.0853 & 0.1168 & 0.2837 & 0.5796 & -0.0028 & 0.0802 \\
Single & STATE & 18.2543 & 0.3333 & 0.2693 & 0.6116 & 0.0004 & 0.5236 & 0.0154 & 0.2386 \\
Single & CellFlow & 1.6758 & 0.0035 & 0.0191 & 0.2860 & 0.7109 & 0.8072 & 0.6138 & 0.6753 \\
Single & scDFM & \underline{1.6186} & \underline{0.0030} & \underline{0.0190} & \underline{0.6957} & \underline{0.7127} & \textbf{0.8914} & \underline{0.6659} & \underline{0.8116} \\
Single & SynGeneFlow (ours) & \textbf{1.5482} & \textbf{0.0026} & \textbf{0.0182} & \textbf{0.7248} & \textbf{0.7426} & \underline{0.8873} & \textbf{0.7045} & \textbf{0.8462} \\
\midrule
Double & Control & 4.1882 & 0.0207 & 0.0423 & N.A. & N.A. & 0.5322 & -0.1303 & -0.0265 \\
Double & scGPT & 3.5171 & 0.0153 & 0.0362 & -0.0665 & 0.5693 & 0.5578 & 0.2814 & 0.2652 \\
Double & GEARS & 3.7458 & 0.0156 & 0.0708 & 0.2543 & 0.7552 & 0.8766 & 0.6407 & 0.8413 \\
Double & Geneformer & 2.0819 & 0.0050 & 0.0237 & 0.3468 & 0.7361 & 0.8067 & 0.6245 & 0.7261 \\
Double & CPA & 5.7891 & 0.0357 & 0.0796 & 0.3652 & 0.4176 & 0.6311 & 0.2432 & 0.2870 \\
Double & STATE & 18.4458 & 0.3404 & 0.2733 & 0.4071 & 0.0061 & 0.5289 & -0.0023 & 0.2580 \\
Double & CellFlow & 2.1042 & 0.0049 & 0.0236 & 0.5074 & 0.8095 & 0.8622 & 0.6780 & 0.7155 \\
Double & scDFM & \underline{2.0309} & \underline{0.0047} & \underline{0.0235} & \underline{0.5676} & \underline{0.8357} & \textbf{0.9189} & \underline{0.7769} & \underline{0.8688} \\
Double & SynGeneFlow (ours) & \textbf{1.8846} & \textbf{0.0041} & \textbf{0.0221} & \textbf{0.6312} & \textbf{0.8624} & \underline{0.9142} & \textbf{0.8136} & \textbf{0.8965} \\
\bottomrule
\end{tabular}%
}
\end{table*}

\subsection{Non-additive Synergy Prediction}
\label{section:synergy_prediction}

To directly evaluate whether SynGeneFlow captures interaction-induced effects rather than only improving full-profile reconstruction, we further assess non-additive synergy prediction on Norman double perturbations. For each double perturbation, we decompose the transcriptional response into an additive component estimated from single-perturbation effects and a residual synergy component that captures non-additive interactions.

\begin{figure}[t]
\centering
\includegraphics[width=\columnwidth]{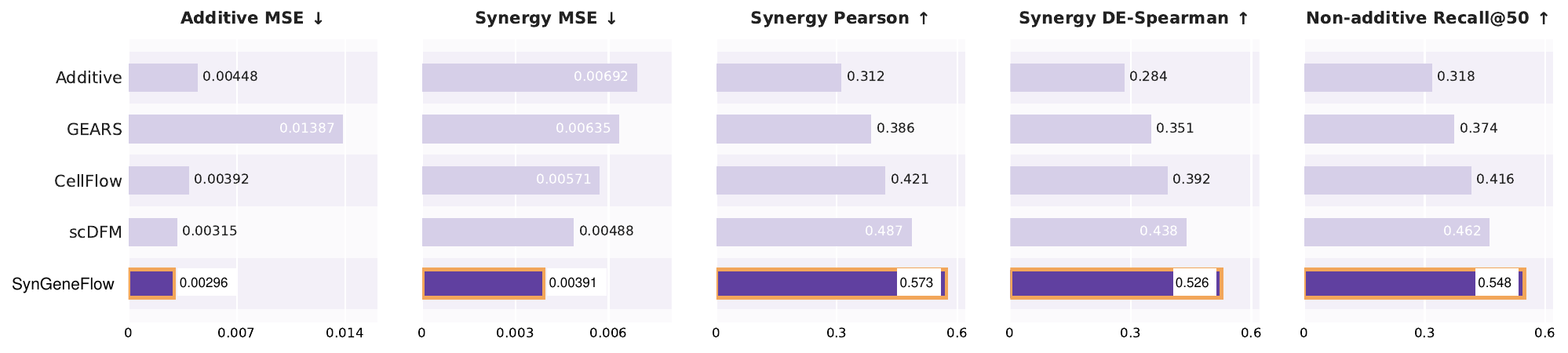}
\caption{
\textbf{Non-additive synergy prediction on Norman double perturbations.}
Each panel reports one evaluation metric using the raw metric values. Lower values are better for Additive MSE and Synergy MSE, while higher values are better for Synergy Pearson, Synergy DE-Spearman $\rho$, and Non-additive Recall@50. SynGeneFlow consistently improves both residual reconstruction and interaction recovery over existing baselines.
}
\label{fig:synergy_prediction}
\end{figure}

Figure~\ref{fig:synergy_prediction} summarizes the results. SynGeneFlow achieves the lowest additive reconstruction error and the lowest synergy reconstruction error, reducing Additive MSE from 0.00315 to 0.00296 and Synergy MSE from 0.00488 to 0.00391 relative to scDFM. More importantly, SynGeneFlow also achieves the strongest residual recovery, improving Synergy Pearson from 0.487 to 0.573, Synergy DE-Spearman $\rho$ from 0.438 to 0.526, and Non-additive Recall@50 from 0.462 to 0.548. These gains indicate that the advantage of SynGeneFlow does not merely come from better endpoint fitting. Instead, the additive-prior residual-flow formulation more effectively isolates and models the non-additive component of combinatorial perturbation responses. This is consistent with the design of SynGeneFlow, where the additive anchor absorbs first-order effects and allows the generative model to focus its capacity on synergy-specific residual structure.

These results highlight that the main benefit of SynGeneFlow lies in modeling the residual structure of combinatorial responses rather than simply improving average reconstruction. In this setting, the additive component captures the dominant first-order response, whereas the residual component contains the interaction-dependent signal that determines whether a model can recover true perturbation synergy. By anchoring generation at the additive estimate, SynGeneFlow reduces the burden of modeling the full transcriptional profile and concentrates the flow on the biologically informative non-additive deviation. This provides a direct explanation for its stronger Synergy Pearson, DE-Spearman $\rho$, and Non-additive Recall@50.

\subsection{Drug Perturbation Transfer}

We further evaluate on ComboSciPlex to test transfer from genetic perturbations to drug perturbations. Unlike Norman, ComboSciPlex measures transcriptional responses to chemical interventions, which can involve broader pathway-level effects and different response structures. This setting tests whether the proposed residual-flow formulation extends beyond CRISPR-based perturbations.

Table~\ref{tab:combosciplex} reports results on ComboSciPlex. CPA is a strong non-flow baseline and achieves the highest DS, while scGPT obtains a competitive Pearson $\Delta$ but negative DE-Spearman $\rho$. scDFM is again a strong direct flow-matching baseline, achieving strong reconstruction and perturbation-response metrics. SynGeneFlow improves over scDFM on L2, MSE, MAE, DE-Spearman $\rho$, and Pearson $\Delta$, reducing MAE from 0.0220 to 0.0207 and increasing DE-Spearman $\rho$ from 0.8289 to 0.8618. Its DS is lower than CPA but higher than scDFM, suggesting that residual-flow modeling improves response recovery while maintaining competitive population separability under drug perturbation transfer. These results indicate that additive-prior residual flow generalizes beyond CRISPR-based perturbations to a distinct chemical perturbation modality.

\begin{table}[t]
\centering
\caption{Performance on ComboSciPlex drug-combination perturbation prediction.}
\label{tab:combosciplex}
\resizebox{\columnwidth}{!}{%
\begin{tabular}{lcccccc}
\toprule
\textbf{Model}
& \textbf{L2 $\downarrow$}
& \textbf{MSE $\downarrow$}
& \textbf{MAE $\downarrow$}
& \textbf{DE-Spearman $\rho$ $\uparrow$}
& \textbf{Pearson $\Delta$ $\uparrow$}
& \textbf{DS $\uparrow$} \\
\midrule
Control & 5.3716 & 0.0324 & 0.0698 & N.A. & N.A. & 0.5714 \\
scGPT & 1.6934 & 0.0031 & 0.0251 & -0.1261 & 0.8322 & 0.8571 \\
CPA & 1.6592 & 0.0029 & 0.0240 & 0.7906 & 0.8150 & \textbf{0.8980} \\
scDFM & \underline{1.6567} & \underline{0.0028} & \underline{0.0220} & \underline{0.8289} & \underline{0.8933} & 0.8776 \\
\midrule
SynGeneFlow (ours) 
& \textbf{1.5814} 
& \textbf{0.0025} 
& \textbf{0.0207} 
& \textbf{0.8618} 
& \textbf{0.9146} 
& \underline{0.8892} \\
\bottomrule
\end{tabular}%
}
\end{table}

\subsection{Ablation Study}
\label{section:ablation}

We perform ablation experiments on the Norman holdout double-perturbation split to examine the main components of SynGeneFlow. This setting stresses both unseen perturbation extrapolation and combinatorial response modeling. Each variant removes or simplifies one component, covering residual flow design, additive-effect supervision, distribution alignment, perturbation-conditioned gene interaction, and velocity backbone design.

\begin{figure}[t]
\centering
\includegraphics[width=\columnwidth]{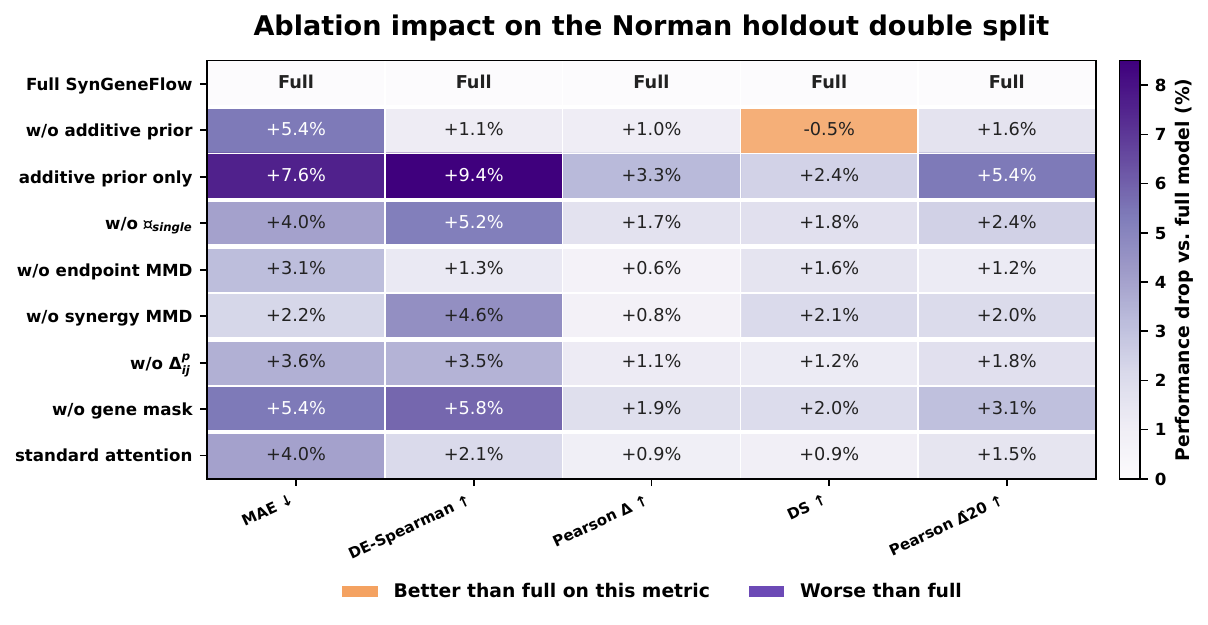}
\caption{
\textbf{Ablation impact on the Norman holdout double split.}
Each cell reports the relative performance change compared with the full SynGeneFlow model. Positive values indicate degradation relative to the full model, while negative values indicate slight improvement. Darker purple denotes a larger performance drop, while orange denotes a slight improvement over the full model on that metric.
}
\label{fig:ablation_impact}
\end{figure}

As shown in Figure~\ref{fig:ablation_impact}, the full model achieves the best MAE, DE-Spearman $\rho$, Pearson $\Delta$, and Pearson $\hat{\Delta}_{20}$. Removing the additive prior increases MAE from 0.0224 to 0.0236 and reduces Pearson $\hat{\Delta}_{20}$ from 0.8794 to 0.8653, showing that direct full-profile transport is weaker than additive-prior residual transport. The additive-prior-only variant further degrades DE-Spearman $\rho$ and Pearson-based metrics, indicating that additive composition captures the dominant response but cannot replace residual velocity modeling. Removing $\mathcal{L}_{\mathrm{single}}$ also hurts performance, suggesting that single-effect supervision constrains the additive operator and prevents it from absorbing higher-order interactions.

Distribution and gene-interaction ablations show similar trends. Removing endpoint MMD or synergy-residual MMD reduces response recovery, with synergy-residual MMD causing a larger drop in DE-Spearman $\rho$ and Pearson $\hat{\Delta}_{20}$. This suggests that residual alignment preserves perturbation-sensitive structure beyond endpoint fidelity. Removing $\Delta_{ij}^{p}$ or the gene mask weakens performance, supporting perturbation-aware gene rewiring within a constrained biological interaction space. Replacing differential attention with standard attention also reduces the main response metrics. Although the variant without the additive prior obtains the highest DS, the full model gives the best overall balance between reconstruction accuracy, perturbation-response recovery, and population-level fidelity.

\section{Conclusion}

We presented \textbf{SynGeneFlow}, a flow matching framework that anchors generation at an additive prior to focus on non-additive perturbation effects. By combining residual flow matching with perturbation-conditioned gene interaction modeling and residual distribution alignment, SynGeneFlow targets interaction-driven responses. Experiments across observed-single, held-out, and drug perturbation settings show that this formulation improves reconstruction and response-direction recovery while preserving distributional fidelity. These results support residual modeling for unseen combinatorial and cross-modality response prediction.

\bibliographystyle{IEEEtran}  
\bibliography{references.bib}      

@article{bunne2023learning,
  title={Learning single-cell perturbation responses using neural optimal transport},
  author={Bunne, Charlotte and Stark, Stefan G and Gut, Gabriele and Del Castillo, Jacobo Sarabia and Levesque, Mitch and Lehmann, Kjong-Van and Pelkmans, Lucas and Krause, Andreas and R{\"a}tsch, Gunnar},
  journal={Nature methods},
  volume={20},
  number={11},
  pages={1759--1768},
  year={2023},
  publisher={Nature Publishing Group US New York}
}

@article{lotfollahi2023predicting,
  title={Predicting cellular responses to complex perturbations in high-throughput screens},
  author={Lotfollahi, Mohammad and Klimovskaia Susmelj, Anna and De Donno, Carlo and Hetzel, Leon and Ji, Yuge and Ibarra, Ignacio L and Srivatsan, Sanjay R and Naghipourfar, Mohsen and Daza, Riza M and Martin, Beth and others},
  journal={Molecular systems biology},
  volume={19},
  number={6},
  pages={MSB202211517},
  year={2023},
  publisher={Springer}
}

@article{qi2024predicting,
  title={Predicting transcriptional responses to novel chemical perturbations using deep generative model for drug discovery},
  author={Qi, Xiaoning and Zhao, Lianhe and Tian, Chenyu and Li, Yueyue and Chen, Zhen-Lin and Huo, Peipei and Chen, Runsheng and Liu, Xiaodong and Wan, Baoping and Yang, Shengyong and others},
  journal={Nature Communications},
  volume={15},
  number={1},
  pages={9256},
  year={2024},
  publisher={Nature Publishing Group UK London}
}

@article{vinas2025systema,
  title={Systema: a framework for evaluating genetic perturbation response prediction beyond systematic variation},
  author={Vi{\~n}as Torn{\'e}, Ramon and Wiatrak, Maciej and Piran, Zoe and Fan, Shuyang and Jiang, Liangze and Teichmann, Sarah A and Nitzan, Mor and Brbi{\'c}, Maria},
  journal={Nature Biotechnology},
  pages={1--10},
  year={2025},
  publisher={Nature Publishing Group US New York}
}

@article{roohani2024predicting,
  title={Predicting transcriptional outcomes of novel multigene perturbations with GEARS},
  author={Roohani, Yusuf and Huang, Kexin and Leskovec, Jure},
  journal={Nature Biotechnology},
  volume={42},
  number={6},
  pages={927--935},
  year={2024},
  publisher={Nature Publishing Group US New York}
}

@article{mathur2022combi,
  title={Combi-seq for multiplexed transcriptome-based profiling of drug combinations using deterministic barcoding in single-cell droplets},
  author={Mathur, Lukas and Szalai, B and Du, NH and Utharala, Ramesh and Ballinger, Martine and Landry, JJM and Ryckelynck, M and Benes, Vladimir and Saez-Rodriguez, Julio and Merten, Christoph A},
  journal={Nature communications},
  volume={13},
  number={1},
  pages={4450},
  year={2022},
  publisher={Nature Publishing Group UK London}
}

@article{ahlmann2025deep,
  title={Deep-learning-based gene perturbation effect prediction does not yet outperform simple linear baselines},
  author={Ahlmann-Eltze, Constantin and Huber, Wolfgang and Anders, Simon},
  journal={Nature Methods},
  volume={22},
  number={8},
  pages={1657--1661},
  year={2025},
  publisher={Nature Publishing Group US New York}
}

@article{csendes2025benchmarking,
  title={Benchmarking foundation cell models for post-perturbation RNA-seq prediction},
  author={Csendes, Gerold and Sanz, Gema and Szalay, Krist{\'o}f Z and Szalai, Bence},
  journal={BMC genomics},
  volume={26},
  number={1},
  pages={393},
  year={2025},
  publisher={Springer}
}

@article{norman2019exploring,
  title={Exploring genetic interaction manifolds constructed from rich single-cell phenotypes},
  author={Norman, Thomas M and Horlbeck, Max A and Replogle, Joseph M and Ge, Alex Y and Xu, Albert and Jost, Marco and Gilbert, Luke A and Weissman, Jonathan S},
  journal={Science},
  volume={365},
  number={6455},
  pages={786--793},
  year={2019},
  publisher={American Association for the Advancement of Science}
}

@article{dai2024gene,
  title={Gene regulatory network inference in the presence of dropouts: a causal view},
  author={Dai, Haoyue and Ng, Ignavier and Luo, Gongxu and Spirtes, Peter and Stojanov, Petar and Zhang, Kun},
  journal={arXiv preprint arXiv:2403.15500},
  year={2024}
}

@article{zhou2025quantifying,
  title={Quantifying batch effects for individual genes in single-cell data},
  author={Zhou, Yang and Sheng, Qiongyu and Wang, Guohua and Xu, Li and Jin, Shuilin},
  journal={Nature Computational Science},
  volume={5},
  number={8},
  pages={612--620},
  year={2025},
  publisher={Nature Publishing Group US New York}
}

@article{xing2025gperturb,
  title={GPerturb: Gaussian process modelling of single-cell perturbation data},
  author={Xing, Hanwen and Yau, Christopher},
  journal={Nature Communications},
  volume={16},
  number={1},
  pages={5423},
  year={2025},
  publisher={Nature Publishing Group UK London}
}

@article{song2025decoding,
  title={Decoding heterogeneous single-cell perturbation responses},
  author={Song, Bicna and Liu, Dingyu and Dai, Weiwei and Mcmyn, Natalie F and Wang, Qingyang and Yang, Dapeng and Krejci, Adam and Vasilyev, Anatoly and Untermoser, Nicole and Loregger, Anke and others},
  journal={Nature cell biology},
  volume={27},
  number={3},
  pages={493--504},
  year={2025},
  publisher={Nature Publishing Group UK London}
}

@article{lipman2022flow,
  title={Flow matching for generative modeling},
  author={Lipman, Yaron and Chen, Ricky TQ and Ben-Hamu, Heli and Nickel, Maximilian and Le, Matt},
  journal={arXiv preprint arXiv:2210.02747},
  year={2022}
}

@article{gretton2006kernel,
  title={A kernel method for the two-sample-problem},
  author={Gretton, Arthur and Borgwardt, Karsten and Rasch, Malte and Sch{\"o}lkopf, Bernhard and Smola, Alex},
  journal={Advances in neural information processing systems},
  volume={19},
  year={2006}
}

@article{adduri2025predicting,
  title={Predicting cellular responses to perturbation across diverse contexts with State},
  author={Adduri, Abhinav K and Gautam, Dhruv and Bevilacqua, Beatrice and Imran, Alishba and Shah, Rohan and Naghipourfar, Mohsen and Teyssier, Noam and Ilango, Rajesh and Nagaraj, Sanjay and Dong, Mingze and others},
  journal={BioRxiv},
  pages={2025--06},
  year={2025},
  publisher={Cold Spring Harbor Laboratory}
}

@article{roohani2025virtual,
  title={Virtual Cell Challenge: Toward a Turing test for the virtual cell},
  author={Roohani, Yusuf H and Hua, Tony J and Tung, Po-Yuan and Bounds, Lexi R and Yu, Feiqiao B and Dobin, Alexander and Teyssier, Noam and Adduri, Abhinav and Woodrow, Alden and Plosky, Brian S and others},
  journal={Cell},
  volume={188},
  number={13},
  pages={3370--3374},
  year={2025},
  publisher={Elsevier}
}

@article{srivatsan2020massively,
  title={Massively multiplex chemical transcriptomics at single-cell resolution},
  author={Srivatsan, Sanjay R and McFaline-Figueroa, Jos{\'e} L and Ramani, Vijay and Saunders, Lauren and Cao, Junyue and Packer, Jonathan and Pliner, Hannah A and Jackson, Dana L and Daza, Riza M and Christiansen, Lena and others},
  journal={Science},
  volume={367},
  number={6473},
  pages={45--51},
  year={2020},
  publisher={American Association for the Advancement of Science}
}

@article{theodoris2023transfer,
  title={Transfer learning enables predictions in network biology},
  author={Theodoris, Christina V and Xiao, Ling and Chopra, Anant and Chaffin, Mark D and Al Sayed, Zeina R and Hill, Matthew C and Mantineo, Helene and Brydon, Elizabeth M and Zeng, Zexian and Liu, X Shirley and others},
  journal={Nature},
  volume={618},
  number={7965},
  pages={616--624},
  year={2023},
  publisher={Nature Publishing Group UK London}
}

@article{yang2022scbert,
  title={scBERT as a large-scale pretrained deep language model for cell type annotation of single-cell RNA-seq data},
  author={Yang, Fan and Wang, Wenchuan and Wang, Fang and Fang, Yuan and Tang, Duyu and Huang, Junzhou and Lu, Hui and Yao, Jianhua},
  journal={Nature machine intelligence},
  volume={4},
  number={10},
  pages={852--866},
  year={2022},
  publisher={Nature Publishing Group UK London}
}

@article{khan2023reusability,
  title={Reusability report: Learning the transcriptional grammar in single-cell RNA-sequencing data using transformers},
  author={Khan, Sumeer Ahmad and Maillo, Alberto and Lagani, Vincenzo and Lehmann, Robert and Kiani, Narsis A and Gomez-Cabrero, David and Tegner, Jesper},
  journal={Nature Machine Intelligence},
  volume={5},
  number={12},
  pages={1437--1446},
  year={2023},
  publisher={Nature Publishing Group UK London}
}

@article{cui2024scgpt,
  title={scGPT: toward building a foundation model for single-cell multi-omics using generative AI},
  author={Cui, Haotian and Wang, Chloe and Maan, Hassaan and Pang, Kuan and Luo, Fengning and Duan, Nan and Wang, Bo},
  journal={Nature methods},
  volume={21},
  number={8},
  pages={1470--1480},
  year={2024},
  publisher={Nature Publishing Group US New York}
}

@article{rosen2023universal,
  title={Universal cell embeddings: A foundation model for cell biology},
  author={Rosen, Yanay and Roohani, Yusuf and Agarwal, Ayush and Samotor{\v{c}}an, Leon and Tabula Sapiens Consortium and Quake, Stephen R and Leskovec, Jure},
  journal={BioRxiv},
  pages={2023--11},
  year={2023},
  publisher={Cold Spring Harbor Laboratory}
}

@article{yu2025perturbnet,
  title={Perturbnet predicts single-cell responses to unseen chemical and genetic perturbations},
  author={Yu, Hengshi and Qian, Weizhou and Song, Yuxuan and Welch, Joshua D},
  journal={Molecular Systems Biology},
  volume={21},
  number={8},
  pages={960},
  year={2025}
}

@article{klein2025cellflow,
  title={CellFlow enables generative single-cell phenotype modeling with flow matching},
  author={Klein, Dominik and Fleck, Jonas Simon and Bobrovskiy, Daniil and Zimmermann, Lea and Becker, S{\"o}ren and Palma, Alessandro and Dony, Leander and Tejada-Lapuerta, Alejandro and Huguet, Guillaume and Lin, Hsiu-Chuan and others},
  journal={bioRxiv},
  pages={2025--04},
  year={2025},
  publisher={Cold Spring Harbor Laboratory}
}

@article{mejia2025diversity,
  title={Diversity by design: Addressing mode collapse improves scrna-seq perturbation modeling on well-calibrated metrics},
  author={Mejia, Gabriel M and Miller, Henry E and Leblanc, Francis JA and Wang, Bo and Swain, Brendan and Camillo, Lucas Paulo de Lima},
  journal={arXiv preprint arXiv:2506.22641},
  year={2025}
}

@article{ramakrishnan2025modeling,
  title={Modeling Gene Expression Distributional Shifts for Unseen Genetic Perturbations},
  author={Ramakrishnan, Kalyan and Hedley, Jonathan G and Qu, Sisi and Dokania, Puneet K and Torr, Philip HS and Prada-Medina, Cesar A and Fauqueur, Julien and Martens, Kaspar},
  journal={arXiv preprint arXiv:2507.02980},
  year={2025}
}

@article{yu2026scdfm,
  title={scdfm: Distributional flow matching model for robust single-cell perturbation prediction},
  author={Yu, Chenglei and Wang, Chuanrui and Liao, Bangyan and Wu, Tailin},
  journal={arXiv preprint arXiv:2602.07103},
  year={2026}
}

@article{xi2026grounding,
  title={Grounding Radiology Report Findings into Medical Image Segmentation},
  author={Xi, Suyang and Hu, Songtao and Wang, Shansong and Ding, Hong and Li, Yuheng and He, Wenfeng and Zhang, Kuo and Del Balzo, Luke and Zhong, Chong and Hu, Mingzhe and others},
  journal={npj Digital Medicine},
  year={2026},
  publisher={Nature Publishing Group UK London}
}

@article{xi2026medlvr,
  title={MedLVR: Latent Visual Reasoning for Reliable Medical Visual Question Answering},
  author={Xi, Suyang and Hu, Songtao and Lai, Yuxiang and Dan, Wangyun and Liu, Yaqi and Wang, Shansong and Yang, Xiaofeng},
  journal={arXiv preprint arXiv:2604.09757},
  year={2026}
}

@article{xi2026radiographic,
  title={A radiographic world model for clinical reasoning and evidence generation},
  author={Xi, Suyang and Hu, Songtao and Wang, Shansong and Safari, Mojtaba and del Balzo, Luke and Karim, Ehsan Ul and Hu, Mingzhe and Zhang, Kuo and Wang, Tonghe and Weichselbaum, Ralph R and others},
  journal={arXiv preprint arXiv:2609.07719},
  year={2026}
}

\vspace{12pt}
\color{red}

\end{document}